\documentclass{article}
\usepackage{ijcai26}

\usepackage{times}
\usepackage{soul}
\usepackage{url}
\usepackage[hidelinks]{hyperref}
\usepackage[utf8]{inputenc}
\usepackage[small]{caption}
\usepackage{graphicx}
\usepackage{amsmath}
\usepackage{amsthm}
\usepackage{booktabs}
\usepackage{algorithm}
\usepackage{algorithmic}
\usepackage[switch]{lineno}
\usepackage{amssymb}

\title{BFHD: Bidirectional Feature Harmonization Decomposition for Heterogeneous Clinical Assessments}

\author{
}

\author{
Yuanhao Zhuo$^1$
\and
Zixi Qin$^1$
\and
Ling Qin$^2$
\And
Wanqing Li$^1$\thanks{Corresponding author.}\\
\affiliations
$^1$University of Wollongong, Wollongong, NSW 2522, Australia \\
$^2$The People's Hospital of Guangxi Zhuang Autonomous Region, Nanning, Guangxi 530021, China\\
\emails
\{yzhuo, wanqing\}@uow.edu.au,
zq181@uowmail.edu.au,
lqin@gxams.org.cn
}

\begin{document}

\maketitle
\footnotetext[1]{Code and supplementary technical materials are available at \url{https://github.com/UOW-AMRL/IJCAI26-BFHD}.}

\footnotetext[2]{%
This manuscript is an extended version of the paper originally published in the Proceedings of IJCAI-ECAI 2026, \textcopyright{} 2026 International Joint Conferences on Artificial Intelligence Organization (IJCAI). This version includes additional supplementary material. The original published version is available at \url{https://doi.org/10.24963/ijcai.2026/784}.
For the formal bibliographic reference, see \cite{zhuo2026bfhd}.
}

\begin{abstract}
    Clinical assessments are often collected using heterogeneous assessment systems across centers and time, leading to records that mix different but related sets of measurements. This motivates harmonization beyond total-score linking. We formulate clinical harmonization at the measurement level as a bidirectional recoverability problem: given paired observations from two assessment systems, the goal is to identify which measurements can be reliably translated in both directions within an application-defined tolerance, while separating non-translatable components. We propose Bidirectional Feature Harmonization Decomposition (BFHD), a feasibility-driven framework that enforces bidirectionally coupled translation and uses feature-wise output gating to produce an explicit decomposition in the original measurement space. Experiments on synthetic data and real clinical assessment pairs show that BFHD achieves broader feasible harmonization coverage and improved subset stability compared to baselines.
\end{abstract}

\section{Introduction}

Clinical assessment of chronic and neuropsychiatric conditions relies on a wide range of rating scales, cognitive tests, and questionnaire-based assessments\cite{apa2022dsm5tr,lui2025cognitive}. In real clinical workflows, however, different centers, cohorts, and time periods often employ different combinations or versions of these assessment systems\cite{villalpando2025comparison}. Even when two assessment systems aim to measure similar clinical domains such as memory or executive function, their measurements (features) typically include both overlapping elements and system-specific components\cite{willcutt2005validity}. As patients move between centers or as protocols evolve, longitudinal records naturally become mixtures of heterogeneous assessment systems rather than observations from a single fixed system\cite{lawrence2017systematic}.

This heterogeneity creates a central challenge for clinical data integration and longitudinal analysis\cite{fortin2017harmonization}. Beyond aligning summary scores, practical harmonization requires understanding how two assessment systems relate at the measurement level: which measurements can be reliably translated across systems, which encode system-specific information, and whether such relationships are consistent in both directions\cite{dorans2007linking,kolen2013test}. These questions are critical for cross-protocol integration, retrospective cohort pooling, and interpretation of heterogeneous clinical records\cite{riley2010meta}.

Existing harmonization practices primarily address this problem through total-score linking and calibration methods, which estimate transformations between total scores or low-dimensional summaries\cite{kolen2013test}. While effective for aligning marginal outcomes, these approaches do not characterize bidirectional recoverability at the measurement level between heterogeneous assessment systems\cite{lord2012applications}. In practice, clinicians and researchers are often left without guidance on which individual measurements can be meaningfully harmonized and which should be treated as non-translatable\cite{hammarstrom2016addressing}. In contrast to feature selection approaches that prioritize features based on their utility for downstream prediction, clinical harmonization requires identifying which measurements can be meaningfully translated across assessments while explicitly preserving system-specific information\cite{molnar2020interpretable}.

In this work, we formalize clinical harmonization at the measurement level as a bidirectional recoverability problem. Given paired observations from two assessment systems, we seek to decompose each system into (i) a harmonizable component consisting of measurements that admit reliable bidirectional translation under a bidirectionally coupled model that enforces synchronous updates across directions, and (ii) a complementary non-translatable component whose information cannot be recovered from the other system. Recoverability is defined as a model-relative, feasibility-based notion, extending classical total-score linking from summary transformations to multivariate, feature-level harmonization.

We introduce Bidirectional Feature Harmonization Decomposition (BFHD), a framework that formulates clinical harmonization as a decomposition of measurements into bidirectionally recoverable and non-translatable components, identified directly in the original measurement spaces. BFHD operationalizes this formulation through a bidirectionally coupled translation model that enforces consistent forward and backward relationships between assessment systems, together with feature-wise gating that determines which measurements must be reconstructed to satisfy cross-system feasibility within a clinically meaningful tolerance\cite{louizos2017learning}. This yields an explicit and interpretable decomposition without task supervision or latent projections.

The contributions of this work are as follows:
\begin{itemize}
    \item We formulate clinical harmonization as a bidirectional recoverability problem at the measurement level, and provide a principled, model-relative definition of harmonizable and non-translatable measurements.
    \item We develop a feasibility-driven, single-operator framework with feature-wise gating that identifies large bidirectionally recoverable subsets directly in the original measurement spaces.
    \item We demonstrate on synthetic and real clinical datasets that BFHD reveals meaningful harmonization structure and supports utilities including cross-protocol integration, temporal deployment under protocol drift, and robustness to test-time missingness.
\end{itemize}

\section{Related Work}

\subsection{Clinical Harmonization and Task-Driven Baselines}

Classical psychometric linking and equating methods, including total-score linking \cite{dorans2007linking,kolen2013test}, equipercentile equating \cite{kolen2013test}, and IRT calibration \cite{cella2007patient,lord2012applications}, map between assessment systems primarily at the level of total scores or low-dimensional traits. They support comparability across protocols but do not identify which individual measurements are bidirectionally recoverable between heterogeneous assessments.

In clinical ML, harmonization is often approached via predictive mappings \cite{fortin2017harmonization,dewey2019deepharmony,dinsdale2021learning} using tree-based models such as feature-wise random forests (FRF) \cite{breiman2001random,breiman2017classification} or gradient boosted trees (XGB) \cite{chen2016xgboost}. These models can yield strong unidirectional prediction but their feature attributions depend on a chosen target
and do not characterize intrinsic bidirectional recoverability. ComBat-style statistical harmonization \cite{johnson2007adjusting,beer2020longitudinal} removes site/batch effects as nuisance variation, which is misaligned with our goal of explicitly retaining non-translatable system-specific content.

\subsection{Multiview Learning and Bidirectional Structure}

Multiview learning relates multiple representations of the same individuals. CCA and its variants \cite{hotelling1935most,bach2005probabilistic,yang2019survey,andrew2013deep} and shared--private latent models such as multiview autoencoders and variational approaches \cite{kingma2013auto} capture cross-view dependence via latent projections or likelihood objectives. However, these methods typically operate in latent spaces and do not yield an explicit account of which observed measurements are recoverable across systems in the original measurement space.

Bidirectional mappings have also been explored in translation settings. Cycle-consistent models couple two directions through closed-loop reconstruction losses \cite{zhu2017unpaired}, while tied-weight autoencoder-style models enforce parameter sharing \cite{hinton2006reducing,bengio2013representation}. We implement a cycle-consistent MLP (CMLP) as a bidirectional mapping baseline without explicit subset selection. To realize strict single-operator coupling in our framework, which enforces shared constraints, we leverage invertible networks from the normalizing flow \cite{dinh2014nice,dinh2016density,kingma2018glow}. A tied-weight MLP (TWMLP) serves as a non-invertible alternative instantiation in our ablations.

\subsection{Gating and Its Relation to Feature Selection}

Feature selection methods typically identify features that improve task performance, commonly via sparsity-inducing penalties such as $L_1$ regularization \cite{tibshirani1996regression,ng2004feature} or via neural gating mechanisms \cite{louizos2017learning,yamada2020feature}. In contrast, our gates are optimized as continuous proxies for membership in a bidirectionally recoverable subset under a shared operator, using feasibility-style reconstruction constraints rather than task supervision. This enables an explicit decomposition into recoverable and non-translatable components aligned with clinical harmonization.

\section{Problem Formulation}
\label{sec:problem}
We consider paired observations from two clinical assessment systems, represented by random vectors $X\in\mathbb{R}^{d_X}$ and $Y\in\mathbb{R}^{d_Y}$. Let $I_X=\{1,\dots,d_X\}$ and $I_Y=\{1,\dots,d_Y\}$ denote feature indices. We seek recoverable subsets of measurements $S_X\subseteq I_X$ and $S_Y\subseteq I_Y$,
whose complements are treated as non-translatable.

\subsection{Bidirectional model class}

Recoverability is defined relative to a bidirectional model class that
reflects the mutual nature of harmonization.
Given paired observations from two assessment systems,
we consider two mappings parameterized by $\Phi$:
\[
f_\Phi : \mathcal{X} \rightarrow \mathcal{Y}, \qquad
g_\Phi : \mathcal{Y} \rightarrow \mathcal{X},
\]

A key requirement is that the two directions are not learned independently. Instead, updates to the mapping from $X$ to $Y$ must simultaneously constrain, and be constrained by, the mapping from $Y$ to $X$. That is, learning the relationship between two assessment systems in one direction necessarily informs the relationship in the reverse direction.

We formalize this requirement by restricting $f_\Phi$ and $g_\Phi$ to be bidirectionally coupled: they are derived from a shared parameterization or structural constraint, such that optimizing one direction induces corresponding updates in the other, rather than allowing two unrelated functions linked only by loss terms.

\subsection{\texorpdfstring{$\varepsilon$}{epsilon}-recoverability}

\paragraph{Conceptual definition}

Noise and heterogeneity in clinical assessments make exact recoverability unrealistic. We therefore define a population-level, approximate notion of recoverability. A measurement $X_i$ is said to be $\varepsilon$-recoverable from $Y$ under a shared bidirectional model parameter $\Phi$ if
\begin{equation}
\mathbb{E}\big[\, d( X_i, g_\Phi(Y)_i ) \,\big] < \varepsilon ,
\label{eq:x-eps}
\end{equation}
where $d(\cdot,\cdot)$ is a feature-wise discrepancy measure and $\varepsilon>0$ specifies an application-level tolerance for clinically acceptable agreement rather than a tuning parameter for predictive accuracy.
Analogously, a measurement $Y_j$ is $\varepsilon$-recoverable if
\begin{equation}
\mathbb{E}\big[\, d( Y_j, f_\Phi(X)_j ) \,\big] < \varepsilon .
\label{eq:y-eps}
\end{equation}

Recoverability is defined relative to a chosen bidirectional model class $\mathcal{M}$, in the same way that classical total-score linking defines equivalence relative to a family of allowable linking functions. Smaller values of $\varepsilon$ correspond to stricter harmonization requirements, while larger values admit broader but looser forms of cross-assessment translation. This definition is population-level and combinatorial: it assumes access to the true expectations and to the exact subsets of features whose recoverability constraints are satisfied.

\paragraph{Practical relaxation}

The population-level conditions in \eqref{eq:x-eps}--\eqref{eq:y-eps} cannot be evaluated directly: expectations are unobserved, and searching over all subsets $(S_X,S_Y)$ is combinatorial. We therefore introduce a lightweight relaxation that preserves the semantics of recoverability while making the problem estimable from data.

Each feature is assigned a gate variable $G_{X,i},G_{Y,j} \in [0,1]$ that acts as a continuous proxy for membership in the recoverable subsets. Only gated features contribute to empirical reconstruction losses, so the optimizer keeps a gate open when the shared bidirectional operator can reconstruct that feature with sufficiently small empirical error. This operationalizes the theoretical $\varepsilon$-recoverability condition using standardized feature-wise reconstruction error, which approximate the magnitude of the deviations appearing in \eqref{eq:x-eps}--\eqref{eq:y-eps}.

Features whose reconstruction errors remain large force their gates toward zero and are treated as non-translatable. This provides a compact and data-driven approximation to the conceptual BFHD formulation while maintaining its intended semantics; the neural instantiation is described in Section~\ref{sec:method}.

\subsection{Feasible decompositions}

A triple \((S_X, S_Y, \Phi)\) is called \(\varepsilon\)-feasible if each selected feature satisfies its recoverability condition:
\begin{equation}
(S_X, S_Y, \Phi) \in \mathcal{F}_\varepsilon
\Longleftrightarrow
\begin{cases}
\mathbb{E}[\, d(X_i, g_\Phi(Y)_i) \,] < \varepsilon, 
& \forall i\in S_X, \\[6pt]
\mathbb{E}[\, d(Y_j, f_\Phi(X)_j) \,] < \varepsilon, 
& \forall j\in S_Y \, .
\end{cases}
\label{eq:feasible}
\end{equation}

This definition does not require exact reconstruction and admits feasible solutions under realistic clinical noise levels.

\subsection{The BFHD optimization problem}

The BFHD problem seeks the largest bidirectionally recoverable subsets:
\begin{align}
\max_{\; S_X \subseteq I_X,\; S_Y \subseteq I_Y,\; \Phi \in \mathcal{M}}
\quad &
w |S_X| + (1-w) |S_Y|
\\[4pt]
\text{subject to} \quad &
(S_X, S_Y, \Phi) \in \mathcal{F}_\varepsilon .
\end{align}
Here $w$ is a fixed user-specified weight rather than an optimization variable. Unless otherwise stated, we use $w=0.5$.

An optimal solution \((S_X^\ast, S_Y^\ast, \Phi^\ast)\) yields the decomposition
\begin{align}
X_{\mathrm{rec}} &= X_{S_X^\ast}, 
& X_{\mathrm{nt}} &= X_{I_X \setminus S_X^\ast}, \\
Y_{\mathrm{rec}} &= Y_{S_Y^\ast}, 
& Y_{\mathrm{nt}} &= Y_{I_Y \setminus S_Y^\ast}.
\end{align}

When multiple maximal feasible solutions exist, BFHD recovers one such solution. In practice, stability across random initializations is used to assess identifiability.

\section{Methodology}
\label{sec:method}

This section presents a practical realization of the BFHD framework. The goal is to construct a shared bidirectional operator for feature-level harmonization that jointly defines the mappings $X \rightarrow Y$ and $Y \rightarrow X$, together with feature-wise gating mechanisms that approximate the subsets appearing in the conceptual formulation in Section~\ref{sec:problem}. The proposed method consists of three components: a shared bidirectional backbone with strict parameter sharing across directions (instantiated with an invertible architecture in our main implementation), feature-wise gates defining relaxed recoverable subsets, and cross-view reconstruction losses that implement the recoverability constraints in a differentiable manner.

\subsection{Architecture Overview}

Given paired observations $(X,Y)$, BFHD constructs two cross-view mappings using a shared bidirectional backbone,
which is instantiated as an invertible neural network (INN) in our main implementation. The $X\rightarrow Y$ and $Y\rightarrow X$ mappings correspond to the forward and inverse passes of the same backbone. For $X\!\rightarrow\!Y$ and $Y\!\rightarrow\!X$, we compute
\begin{equation}
\tilde{Y} = f_{\Phi}(X), \qquad 
\tilde{X} = g_{\Phi}(Y),
\label{eq:tilde-maps}
\end{equation}
and apply feature-wise $L_0$ gates at the outputs:
\begin{equation}
\hat{Y}_j = G_{Y,j}\,\tilde{Y}_j, \qquad 
\hat{X}_i = G_{X,i}\,\tilde{X}_i .
\label{eq:gated-outputs}
\end{equation}

The invertible backbone enforces strict bidirectional parameter sharing, while the feature-wise gates determine which measurements in each system are required to be reconstructed. Only gated outputs contribute to the loss, providing a differentiable approximation to the recoverability constraints in the BFHD formulation.

\subsection{Bidirectional Backbone via Invertible Networks}

A key requirement in the BFHD framework is that the mappings $X \rightarrow Y$ and $Y \rightarrow X$ share a common parameterization. To satisfy this requirement, we instantiate $(f_{\Phi}, g_{\Phi})$ using an invertible neural architecture. Unlike conventional flow-based models, the invertible structure is not used for likelihood estimation or sampling; instead, it provides a single coupled operator that guarantees strict bidirectional parameter sharing.

\paragraph{Instantiation with a NICE-style architecture}

We adopt a NICE-style architecture because its additive coupling provides stable training for tabular, low-dimensional clinical data and avoids the scaling instability of affine coupling, which is unnecessary in our non-likelihood setting\cite{dinh2014nice}. 

Because the NICE architecture is exactly invertible by construction, the two cross-view mappings used in BFHD correspond to a single operator $F_{\Phi}$ and its analytic inverse $F_{\Phi}^{-1}$. This guarantees strict bidirectional parameter sharing and thus directly satisfies the requirement specified in the BFHD formulation. Importantly, the invertible structure serves not as a generative modeling assumption, but as a mechanism that enforces a tightly coupled forward--backward relationship between the two clinical systems.

\paragraph{Dimension alignment}

When $d_X \neq d_Y$, both features are embedded into a common dimension $D = \max(d_X, d_Y)$ by zero-padding the lower-dimensional system. Padding dimensions are never supervised in the reconstruction losses and do not interact with the gating mechanism. This ensures that the decomposition concerns only the original features.

\paragraph{Role of Invertibility in BFHD}
Invertibility is used solely to enforce strict parameter sharing without assuming the invertibility in clinical assessment systems. The X→Y and Y→X mappings are the forward and inverse passes of a single operator. This eliminates forward–backward drift common in cycle-consistent dual networks and improves subset stability. After output gating, the overall mapping is intentionally non-invertible. We use a NICE-style INN as a convenient instantiation; a non-invertible tied-weight backbone is evaluated in ablations.

\subsection{Output $L_0$ Gating}

The BFHD problem requires determining, for each feature of $X$ and $Y$, whether it belongs to the bidirectionally recoverable subset. To obtain a differentiable relaxation of these binary decisions, we introduce feature-wise gates
\[
G_X = (G_{X,1}, \ldots, G_{X,d_X}), 
\qquad
G_Y = (G_{Y,1}, \ldots, G_{Y,d_Y}),
\]
with values in $[0,1]$. A gate value near $1$ indicates that the corresponding feature is treated as recoverable, and a value near $0$ indicates that the feature is assigned to the non-translatable component.

Unlike conventional feature selection methods that gate inputs, BFHD applies gates at the outputs of the cross-view mappings. Thus the predicted vectors are masked according to the definitions in Eqs.~\eqref{eq:gated-outputs}, ensuring that the model always receives the full input $X$ or $Y$ while only the gated subset is required to be reconstructed. This matches the recoverability semantics of the BFHD formulation.

The gates are implemented using the hard-concrete relaxation of the $L_0$ penalty \cite{louizos2017learning}. This parameterization yields stochastic binary gate variables during training while remaining differentiable in expectation, allowing gradient-based optimization of feature inclusion.

\subsection{Gated Hinge Losses}
\label{sec:gated-hinge}

The BFHD formulation defines $\varepsilon$-recoverability through measurement-level reconstruction error tolerances (Section~\ref{sec:problem}). To approximate this conceptual feasibility constraint from data, we introduce a gated hinge loss with a stricter training tolerance $\varepsilon_{\text{train}}$, which serves as a conservative proxy for the target evaluation tolerance.

We first define a scale-normalized reconstruction error $\mathrm{err}(\cdot,\cdot)$ for individual features. In our implementation, we use feature-wise normalized mean squared error (NMSE). Normalization ensures that $\varepsilon_{\text{train}}$ has a consistent interpretation across heterogeneous features. Throughout this work, all assessment features are treated as numerical variables, and the same normalized squared-error-based discrepancy is used for all features. Ordinal items, when present, are modeled as numerical scores under this approximation.

Using $\mathrm{err}$, we define the gated hinge losses for the two cross-view directions:
\begin{equation}
\mathcal{L}_{X \leftarrow Y}
=
\sum_{i=1}^{d_X}
G_{X,i}\,
\mathrm{ReLU}\bigl(\mathrm{err}(X_i,\tilde{X}_i)-\varepsilon_{\text{train}}\bigr),
\label{eq:hinge-x}
\end{equation}
\begin{equation}
\mathcal{L}_{Y \leftarrow X}
=
\sum_{j=1}^{d_Y}
G_{Y,j}\,
\mathrm{ReLU}\bigl(\mathrm{err}(Y_j,\tilde{Y}_j)-\varepsilon_{\text{train}}\bigr).
\label{eq:hinge-y}
\end{equation}

This loss has a direct feasibility interpretation. A gated feature incurs no penalty once its reconstruction error falls below $\varepsilon_{\text{train}}$, and therefore does not encourage over-optimization beyond the feasibility threshold. In practice, $\varepsilon_{\text{train}}$ is set substantially smaller than the evaluation tolerance $\varepsilon_{\text{cutoff}}$ to discourage borderline features from remaining gated open due to noise or finite-sample effects, thereby improving subset stability across random initializations. Penalties are applied only when the underlying $\varepsilon$-recoverability constraint cannot be satisfied under the shared bidirectional operator, creating pressure to either improve reconstruction or reduce the corresponding gate value.

\subsection{Gate Reward Term}

Using the same hard-concrete parameterization as above, we optimize gates to favor inclusion rather than sparsity. Unlike standard $L_0$ regularization, which penalizes gate activation to promote sparsity, BFHD requires the opposite behavior. Gates should remain open unless cross-view reconstruction forces them to close. Accordingly, we use a reward rather than a penalty:
\[
\mathcal{L}_{\text{gate}}
=
-\lambda_{L_0}
\left(
\sum_{i=1}^{d_X} G_{X,i}
+
\sum_{j=1}^{d_Y} G_{Y,j}
\right),
\qquad
\lambda_{L_0} > 0.
\]

This term broadens the candidate recoverable subsets, while the gated cross-view losses apply opposing pressure by closing gates that lead to poor bidirectional reconstruction. Their interaction forms a continuous relaxation of the BFHD objective and identifies the largest empirically recoverable subsets.

\subsection{Overall Objective}

The complete training objective combines the gated cross-view reconstruction terms with the gate reward term:
\[
\mathcal{L}_{\text{total}}
=
\mathcal{L}_{X \leftarrow Y}
+
\mathcal{L}_{Y \leftarrow X}
+
\mathcal{L}_{\text{gate}}.
\]
Minimizing $\mathcal{L}_{\text{total}}$ jointly optimizes the shared bidirectional operator and the gating variables, producing an empirical approximation to the BFHD decomposition.

Training is performed using a staged procedure to stabilize optimization and avoid degenerate solutions; details in Appendix~\ref{sec:sm-training}.

\section{Experiments}

\subsection{Experimental Setup}

\paragraph{Datasets:}
We use (1) a synthetic dataset generated using a known shared/private/noise-factor model (details and exact parameters in Appendix~\ref{sec:sm-synth-params}). This provides ground-truth recoverable and non-translatable measurement labels for evaluating structure recovery. (2) Alzheimer’s Disease Neuroimaging Initiative (ADNI) cognitive assessments\cite{mueller2005alzheimer}, and (3) a paired real-world ADHD assessment dataset consisting of two widely used symptom scales,
collected under a controlled co-administration protocol and available upon reasonable request.  All real-world datasets consist of paired, co-administered assessment systems with partially overlapping but non-identical measurement sets. To avoid subject-level leakage, each subject contributes at most one observation in all experiments unless otherwise stated.

\paragraph{Baselines:}
We compare our approach with: 
\begin{enumerate}
    \item Clinically-used harmonization and predictive baselines (IRT, FRF, XGB), which learn unidirectional or task-driven mappings;
    \item Bidirectional learning models (CMLP, NICE) that enforce forward-backward consistency without explicit feasibility-based gating;
    \item Ablated variants of our approach.
\end{enumerate}

\paragraph{Implementation Details.}
Implementation details, model architectures, and training hyperparameters are reported in Appendix~\ref{sec:sm-impl}. All experiments use 10 random seeds to assess stability ($s_{train}\in\{1,...,10\}$), and synthetic datasets are generated using fixed data-generation seeds ($s_{gen}\in\{1,...,5\}$). We distinguish between two tolerances: a training tolerance $\varepsilon_{\text{train}}$, used only for optimizing BFHD, and an evaluation tolerance $\varepsilon_{\text{cutoff}}$, used uniformly across all methods for reporting feasible measurements.

The feasibility tolerance $\varepsilon_{\text{cutoff}}$ is system- and context-dependent rather than universal. Prior psychometric analyses report substantial intrinsic measurement uncertainty for NeuroBAT subtests in ADNI \cite{crane2023measurement} and SNAP-IV items in ADHD \cite{huang2023differences}. When measurement-level variance estimates are available, these uncertainties can be mapped to the normalized error scale used in our experiments; otherwise, they anchor the order of magnitude of acceptable reconstruction error. We therefore adopt $\varepsilon_{\text{cutoff}} = 0.6$ as a conservative, measurement-anchored reference operating point across all experiments, with full protocol details and sensitivity analyses provided in Appendices~\ref{sec:sm-tolerance} and~\ref{sec:sm-exp2}.

\subsection{Experiment 1: Synthetic Feasibility Recovery under Ground-Truth $\varepsilon$}

We first evaluate BFHD in a controlled synthetic setting where ground-truth recoverability labels are available by construction. This experiment focuses on feasibility structure recovery: whether a method can correctly identify which measurements satisfy the $\varepsilon$-recoverability condition, rather than comparing reconstruction accuracy.

\paragraph{Evaluation protocol.}
Recoverability is evaluated as a structure recovery problem, with ground-truth defined at the feature-index (measurement) level. For BFHD, recoverability is determined directly by the learned output gates after binarization. For baseline methods that do not perform gating, recoverability is inferred post hoc using a relaxed feasibility criterion: a feature is classified as recoverable if its reconstruction error falls below $2\varepsilon_{\text{gen}}$, where $\varepsilon_{\text{gen}}$ is the noise tolerance used in synthetic data generation. This conservative relaxation compensates for finite-sample estimation error and favors baseline models that do not explicitly optimize feasibility.

We restrict baselines in this experiment to models that naturally admit bidirectional reconstruction and feature-level feasibility evaluation. Other architectures evaluated in real-world experiments are omitted here, as the synthetic data are designed to probe bidirectional feasibility recovery rather than predictive accuracy under task-specific assumptions.

We report structure recovery accuracy ($Acc$), defined as the proportion of feature indices whose recoverability status is correctly identified, separately for $X$ and $Y$. To assess robustness, we additionally report the stability of the recovered measurement subsets across random seeds using Jaccard ($J$) similarity. Other diagnostic metrics are reported in Appendix~\ref{sec:sm-exp1}.

\paragraph{Results.}
\begin{table}[h]
\centering
\begin{tabular}{lcccc}
\hline
Model
& Acc$_X$
& Acc$_Y$
& $J_X$ 
& $J_Y$\\
\hline
Ours & $0.880_{\pm0.024}$  & $0.830_{\pm0.024}$ & 0.933 
  & 0.924  \\
CMLP  & $0.855_{\pm0.015}$ & $0.810_{\pm0.049}$ & 0.975 
  & 0.789 \\
NICE    & $0.735_{\pm0.039}$  & $0.725_{\pm0.040}$ & 0.837 
  & 0.814 \\
\hline

\end{tabular}
\caption{Synthetic feasibility recovery under ground-truth $\varepsilon$ ($s_{gen}$=5). Acc$_X$ and Acc$_Y$ denote index-level structure recovery accuracy for systems $X$ and $Y$, respectively (mean $_{\pm std}$ across 10 seeds). $J_X$ and $J_Y$ report Jaccard similarity. For baseline models without gating, recoverability is inferred post hoc using a relaxed $2\varepsilon_{\text{gen}}$ feasibility criterion. Additional results are provided in Appendix~\ref{sec:sm-exp1}.}
\label{tab:synthetic-structure}
\end{table}

Table~\ref{tab:synthetic-structure} summarizes the results. Our approach achieves the best overall structure recovery accuracy with strong subset stability across seeds among bidirectional baselines. In particular, methods without explicit feasibility-driven gating exhibit unstable behavior on weakly recoverable measurements near the $2\varepsilon_{\text{gen}}$ boundary, leading to increased variability across random seeds.

\subsection{Experiment 2: Real-World Harmonization}

We next evaluate BFHD on real clinical assessment systems to assess its practical utility for cross-protocol harmonization. In contrast to Experiment~1, where ground-truth recoverability is known, real-world datasets do not admit a definitive notion of measurement-level equivalence. Accordingly, the goal of this experiment is not to maximize reconstruction accuracy, but to characterize how much of each assessment system can be harmonized under a shared bidirectional model while maintaining feasibility and stability.

\paragraph{Evaluation protocol.}
For each paired assessment system and each mapping direction, we evaluate harmonization outcomes using one primary and three supporting metrics. The primary metric is the number of recoverable measurements (\#Rec). For BFHD, recoverable measurements are reported using the learned output gates. For baseline methods, which do not produce an explicit measurement-level recoverability indicator, recoverability is inferred at the feature level using the feasibility criterion $err < \varepsilon_{\text{cutoff}}$. The Jaccard similarity ($J$) of the recovered measurement subsets across random seeds, which quantify harmonization coverage and stability, respectively. Diagnostic metrics summarize the strength of agreement among the selected measurements using concordance correlation coefficient (CCC) thresholds\cite{lawrence1989concordance,barnhart2007overview}. Specifically, we report the counts of selected measurements with strong agreement ($CCC>0.7$) and marginal/moderate agreement ($CCC>0.5$). These CCC-based counts are not used to define feasibility; rather, they help interpret whether the additional measurements included under $\epsilon$-feasibility correspond to clinically meaningful agreement levels.

Unless otherwise stated, all real-world results are reported at a fixed feasibility threshold $\varepsilon_{\text{cutoff}}=0.6$. We further analyze sensitivity to this choice through an $\varepsilon$-sweep in Figure~\ref{fig:epsilon-sweep-embic-neurobat} (with additional settings reported in Appendix~\ref{sec:sm-tolerance}).

\paragraph{Results.}

\begin{table*}[h]
\centering
\begin{tabular}{l l c c c c c c c c c}
\hline
Data & Method & $\#Rec_A$  & $J_A$~& $\# A>0.7~$ & $\#A>0.5~$ & $\#Rec_B$ & $J_B$ & $\# B>0.7~$&$\#B>0.5~$ \\
\hline
                    & IRT & 11 & 0.733 & 4.5 & 10.5 & 7 & 0.785 & 4 & 7\\
CAS(A)              & XGB & 9 & 0.710 & 6 & 9 & 7 & 0.842 & 5 & 7\\
$\leftrightarrow$   & FRF & 9 & 0.683 & 4.5 & 9 & 7.5 & 0.858 & 6 & 7.5\\
SNAP(B)             & CMLP & 9 & 0.640 & 5 & 9 & 7 & 0.764 & 6 & 7\\
                    & NICE & 11 & 0.867 & 4 & 11 & 9 & 0.774 & 5 & 9\\
                    & Ours & 11 & 0.930 & 4 & 11 & 9 & 0.817 & 5 & 9 \\
\hline
                    & IRT & 6 & 0.841 & 1 & 6 & 0 & N/A & 0 & 0\\
EMBIC(A)            & XGB & 6 & 0.806 & 1 & 6 & 1 & 0.4 & 0 & 1\\
$\leftrightarrow$   & FRF & 6 & 0.819 & 1 & 6 & 0 & 0.533 & 0 & 0\\
NeuroBAT            & CMLP & 5 & 0.739 & 1 & 5 & 0 & 0.491 & 0 & 0\\
(B)                 & NICE & 7 & 0.806 & 1 & 7 & 2 & 0.523 & 0 & 1.5\\
                    & Ours & 7 & 0.910 & 1 & 7 & 2 & 0.689 & 0 & 2 \\
\hline
\end{tabular}
\caption{
Real-world harmonization results under $\varepsilon_{\text{cutoff}}=0.6$. For each row, A and B correspond to the two assessment systems. $\#\mathrm{Rec}_A$/$\#\mathrm{Rec}_B$ are the median numbers of recoverable measurements and $J_A$/$J_B$ are Jaccard stability across 10 seeds. $\#A{>}0.7$, $\#A{>}0.5$ (and analogously for B) count selected measurements whose CCC exceeds 0.7 or 0.5, used as diagnostic measures of feasibility.
}
\label{tab:real-harmonization}
\end{table*}

Table~\ref{tab:real-harmonization} summarizes the harmonization results across all dataset pairs. Across all settings, our approach consistently identifies larger recoverable subsets than baseline methods while maintaining comparable or improved stability across random seeds. In particular, our approach recovers substantially more measurements than traditional linking and predictive baselines, reflecting its objective of maximizing feasible harmonization coverage rather than optimizing pointwise reconstruction accuracy.

Importantly, the additional measurements identified by our approach are not dominated by poorly reconstructed measurements. Across datasets, our approach matches baselines on the number of strongly aligned measurements while consistently increasing the number and stability of marginally aligned measurements. This pattern supports our goal of expanding feasible harmonization coverage without sacrificing the clinically reliable core.

\paragraph{Effect of the feasibility tolerance.}
\begin{figure}[t]
    \centering
    \includegraphics[width=\linewidth]{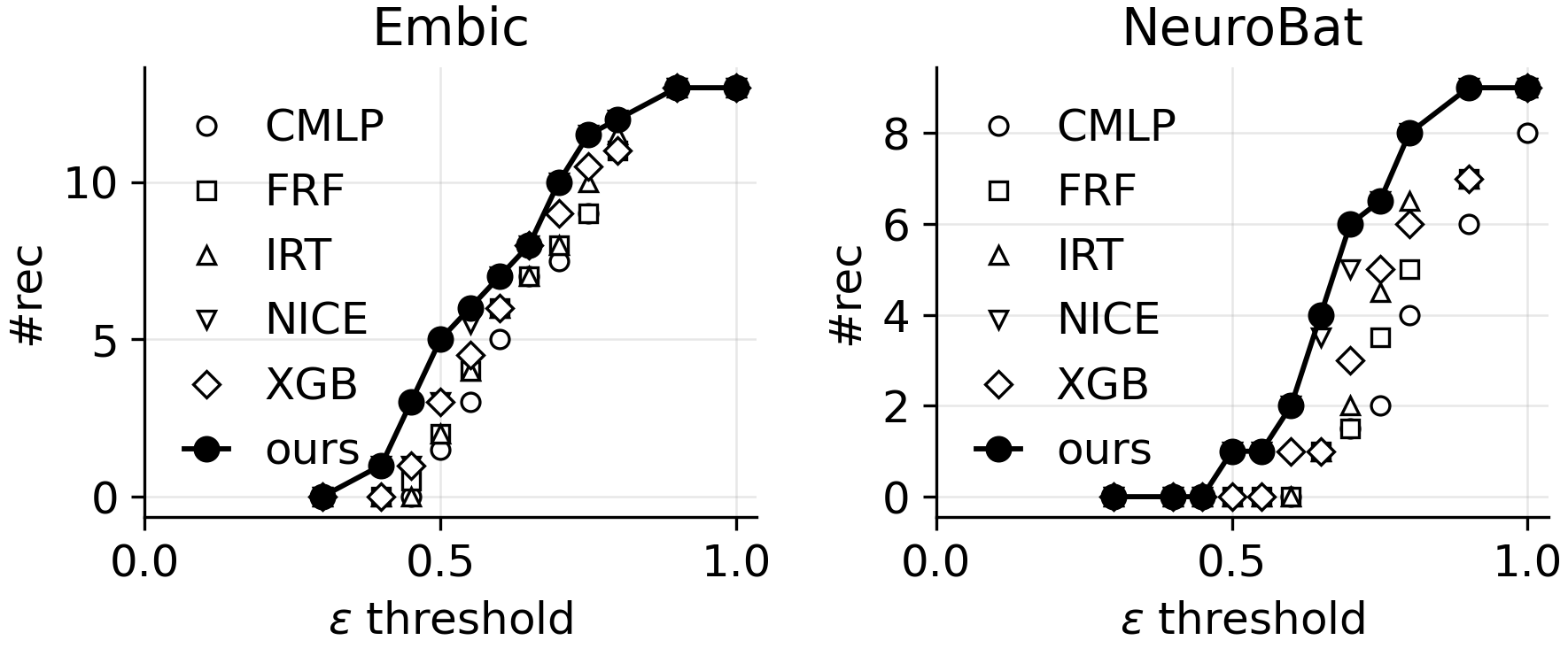}
    \caption{
    $\varepsilon$-sweep on EMBIC and NeuroBAT.
    Median number of recoverable measurements  ($\#\mathrm{rec}$) as a function of the feasibility threshold $\varepsilon$.
    CAS$\leftrightarrow$SNAP results are reported in Appendix~\ref{sec:sm-exp2}.
    }
    \label{fig:epsilon-sweep-embic-neurobat}
\end{figure}

To analyze how harmonization coverage varies with the feasibility criterion, we sweep the tolerance $\varepsilon$ and report the number of recoverable measurements ($\#\mathrm{rec}$).
Figure~\ref{fig:epsilon-sweep-embic-neurobat} shows the resulting curves for the EMBIC$\leftrightarrow$NeuroBAT pair, plotted separately for each assessment system.
Across a broad range of thresholds, our approach consistently recovers more measurements than all baseline methods, demonstrating that it can achieve wider harmonization coverage while respecting the same feasibility constraint.
This advantage is especially pronounced in the intermediate regime (approximately $\varepsilon_{\text{cutoff}}\in[0.55,0.8]$), which corresponds to realistic clinical settings where assessments share a partially overlapping core but differ in system-specific content.
In contrast, baseline methods tend to be either overly conservative at lower thresholds or exhibit abrupt and unstable increases in $\#\mathrm{rec}$ as $\varepsilon_{\text{cutoff}}$ is relaxed.
These results highlight BFHD’s ability to trace a smooth and interpretable coverage--feasibility frontier.
Results for CAS$\leftrightarrow$SNAP and additional dataset pairs are deferred to Appendix~\ref{sec:sm-exp2} as they exhibit similar but less pronounced trends.

\subsection{Experiment 3: Temporal Deployment Utility under Protocol Drift}

While Experiment~2 evaluates harmonization outcomes under random splits from a fixed data distribution, real-world deployment requires applying harmonization decisions learned on earlier cohorts to data collected at a later time. Such temporal shifts may arise from changes in assessment procedures, rater behavior, cohort composition, or missingness patterns. Here we evaluate the temporal deployment utility of BFHD: whether measurement subsets identified as recoverable on early data remain feasible when deployed on later cohorts under potential protocol drift.

\paragraph{Temporal split and evaluation protocol.}
For each real-world dataset, we evaluate temporal deployment utility using a time-based split, where models are trained on early data and evaluated on a held-out later cohort; detailed split procedures and implementation details are provided in Appendix~\ref{sec:sm-split}.

Recoverable measurement subsets are defined on the training split and are not re-selected on the test split. For our approach, recoverable measurements are obtained by binarizing the learned output gates, while baseline methods infer recoverability on the training split using the same feasibility criterion as in Experiment~2. A training-period recoverable measurement is considered retained if it continues to satisfy $\mathrm{err}_{\mathrm{test}} < \varepsilon_{\text{cutoff}}$ on the held-out test split.

We report two deployment utility metrics:
(i) the number of retained recoverable measurements on the test split (\#Ret), and
(ii) the retention rate (Ret\%), defined as the proportion of training-period recoverable measurements that remain feasible at test time.
Agreement strength of retained measurements is reported in Appendix~\ref{sec:sm-exp3}.

\paragraph{Results.}
\begin{table}[t]
\centering
\begin{tabular}{l c c c c}
\hline
Method 
& \#Ret$_A$ 
& Ret\%$_A$
& \#Ret$_B$ 
& Ret\%$_B$ \\
\hline
IRT      &3 &   $42.9_{\pm0.0}$&   0&   N/A\\
XGB      &6 &   $85.7_{\pm0.0}$& 0.5&  $62.5_{\pm48.4}$ \\
FRF      &5 &   $56.0_{\pm8.0}$&   0&  $ 0.0_{\pm0.0}$\\
CMLP   &6 &   $71.0_{\pm7.5}$&   1&  $90.0_{\pm10.0}$\\
NICE     &6 &   $77.4_{\pm9.9}$&   2&  $49.2_{\pm10.5}$\\
Ours     &7 &   $100.0_{\pm0.0}$&  3&  $100.0_{\pm0.0}$\\
\hline
\end{tabular}
\caption{
Temporal deployment utility under protocol drift on the EMBIC(A)$\leftrightarrow$NeuroBAT(B) assessment pair. $\varepsilon_{\text{cutoff}}=0.6$. \#Ret$_A$/\#Ret$_B$ denote the numbers of training-period recoverable measurements that remain feasible on the held-out test split.
Ret\%$_A$/Ret\%$_B$ report the corresponding retention rates. (mean $_{\pm std}$ across 10 seeds)
Additional results in Appendix~\ref{sec:sm-exp3}.
}
\label{tab:temporal-utility}
\end{table}

Table~\ref{tab:temporal-utility} summarizes temporal deployment results for the representative EMBIC$\leftrightarrow$NeuroBAT assessment pair.
Across methods, our approach retains a larger number of training-period recoverable measurements under temporal shift while maintaining higher retention rates than baseline approaches.
These results indicate that BFHD produces harmonization decisions that generalize more reliably across time and are therefore better suited for deployment in longitudinal and evolving clinical settings.

\paragraph{Additional utility analyses.}
Robustness under MCAR missingness\cite{little2019statistical,rubin1976inference} and downstream transfer for prediction tasks are reported in Appendix~\ref{sec:sm-utility}.

\subsection{Experiment 4: Ablation and Stability Analysis}

We examine how architectural and objective components contribute to the stability and behavior of BFHD.

\paragraph{Ablation settings.}
Starting from the full BFHD framework, we consider four ablations:
(i) replacing the bidirectionally coupled invertible NICE backbone with a non-invertible tied weight backbone (TWMLP);
(ii) switching the $L_0$ gate reward to a standard $L_0$ penalty;
(iii) removing output gating; and
(iv) replacing the hinge-based feasibility loss with a standard reconstruction loss.
Each ablation modifies one component while keeping the others fixed.

\paragraph{Evaluation protocol.}
All variants are evaluated on representative real-world system pairs used in Experiment~2.
We report the median number of recoverable measurements (\#Rec) and the Jaccard similarity ($J$) across random seeds, which jointly capture coverage and stability.

\paragraph{Results and observations.}
\begin{table}[t]
\centering
\begin{tabular}{l c c c c}
\hline
Variant 
& $\#\mathrm{Rec}_A$
& $J_A$
& $\#\mathrm{Rec}_B$
& $J_B$\\
\hline
Ours (full)          & 7 & 0.910 & 2 & 0.689 \\
\hline
- TWMLP   & 7 & 0.886 & 1 & 0.600 \\
- $L_{0}$ penalty      & $0^{\dagger}$ & N/A & $0^{\dagger}$ & N/A \\
- No gating        & 7 & 0.806 & 2 & 0.523 \\
- No hinge         & $13^{\dagger}$ & N/A & $9^{\dagger}$ & N/A \\
\hline
\end{tabular}
\caption{
Ablation study on EMBIC (A) $\leftrightarrow$ NeuroBAT (B) under $\varepsilon_{\text{cutoff}}=0.6$.
Notation are consistent with Table~\ref{tab:real-harmonization}.
Additional results in Appendix~\ref{sec:sm-exp4-cas}.
$\dagger$ indicate trivial gating solutions.
}
\label{tab:ablation}

\end{table}

Table~\ref{tab:ablation} summarizes the ablation results. Replacing the $L_0$ gate reward with a standard $L_0$ penalty consistently leads to degenerate solutions, confirming the necessity of explicitly favoring maximal feasible subsets. Removing output gating reduces subset stability, indicating that enforcing feasibility at the reconstruction level is critical for consistent measurement identification. Replacing the invertible backbone with a non-invertible tied-weight MLP preserves recoverable subset size, with dataset-dependent effects on stability. This suggests that strict bidirectional parameter sharing contributes to robustness, while invertibility itself is not a prerequisite for feasibility-based coverage.

Overall, these results show that effective BFHD instantiations depend critically on the joint use of a shared bidirectional backbone, output-level gating, and gated hinge losses.

\section{Conclusion}
We formulated clinical harmonization as a bidirectional recoverability problem and proposed BFHD, a feasibility-driven framework that identifies large bidirectionally recoverable measurement subsets under a shared operator. Across synthetic data and real clinical assessment pairs, BFHD consistently yields broader feasible harmonization coverage and higher subset stability than baselines. BFHD provides an explicit and interpretable decomposition for cross-protocol integration of heterogeneous clinical assessments.

\appendix

\section{Implementation Details}
\label{sec:sm-impl}

\subsection{Backbone Architecture}

BFHD instantiates the shared bidirectional operator using a NICE-style invertible neural network with additive coupling layers. The invertible backbone is used purely as a mechanism for strict parameter sharing across directions, rather than for likelihood modeling.

Each coupling block is parameterized by a multi-layer perceptron with hidden widths $(64,64,32,32)$ for both sides, selected via greedy search due to the large architectural search space of invertible models. Once selected, the same backbone architecture is used across all datasets; only feasibility-related hyperparameters are tuned per dataset.

\subsection{Training Procedure}
\label{sec:sm-training}

Training is performed using a three-stage procedure designed to stabilize joint optimization of the backbone and gating variables.

\paragraph{Stage 1: Backbone warm-up}
The invertible backbone is first trained with all gates forced open. The objective consists of feature-wise reconstruction losses in both directions. This stage provides a reasonable initialization before introducing feasibility constraints.

\paragraph{Stage 2: Joint optimization with feasibility constraints}
In the second stage, backbone parameters and gates are optimized jointly. We employ a hinge-style feasibility loss together with a gate reward implemented in an equivalent close-penalty form. For each feature, the objective takes the form
\[
\lambda_{\text{close}}(1-\bar g)
\;+\;
\bar g \cdot \mathrm{ReLU}\bigl(\mathrm{NMSE}-\varepsilon_{\text{train}}\bigr),
\]
where $\bar g$ denotes the minibatch-averaged gate value. Minimizing the first term is equivalent to maximizing a constant reward for opening the gate, while the second term enforces feasibility. This formulation is numerically stable and equivalent to the gate-reward objective described in the main paper up to a constant shift.

During Stage~2, the training tolerance $\varepsilon_{\text{train}}$ is annealed linearly from an initially loose value to a stricter target value during the gate warm-up phase. In addition, the learning rate for gate parameters follows a staged schedule: it is initialized to a small value during early warm-up ($1\times10^{-5}$), increased during the main joint optimization phase to enable effective output selection ($1\times10^{-3}$), and gradually decreased again to stabilize the final gate configuration ($5\times10^{-4}$). This schedule reduces sensitivity to initialization and improves subset stability.

\paragraph{Stage 3: Gate hardening and fine-tuning}
After Stage~2, gates are binarized and frozen. The backbone is then fine-tuned under the target feasibility tolerance. Early stopping with validation monitoring is applied in all stages, and the best checkpoint is carried forward between stages.

\subsection{Hard-concrete $L_0$ output gates}
\label{sec:sm-l0-gates}

Feature-wise output gates are implemented using a hard-concrete ($L_0$) relaxation.
Each gate is parameterized by a learnable logit and produces a continuous value in $[0,1]$ during training,
which is optionally hardened to a binary mask after convergence.

Unless otherwise stated, we use the following gate parameterization in all experiments:
the Concrete temperature is set to $1/8$;
the stretched interval is fixed to $[-1,\,2]$;
and gate logits are initialized to match an initial drop rate of $0.65$.
These values are kept fixed across datasets and random seeds.

After training, gates are binarized and frozen for backbone fine-tuning.
A measurement is reported as recoverable if its hardened gate is active.

\subsection{Optimization and Learning Rates}

For all gated models (Ours and TWMLP), we optimize backbone parameters and gate parameters using separate learning rates. The backbone learning rate is fixed to $5\times10^{-5}$ during stage 2 across all experiments, while gate parameters use a higher learning rate shared across gated models. This separation is critical in practice:
using a single learning rate leads to unstable gating behavior or degraded backbone mappings.

All optimizations for neural network based models use Adam. Early stopping based on validation performance is applied to all neural network based models, including NICE, ours, TWMLP, and CMLP. Baselines without gating (CMLP, IRT, FRF, XGB) are trained using a single optimizer and a single learning rate.

\subsection{Hyperparameter Selection}

The hyperparameter space of invertible networks is large,
so we do not perform exhaustive grid search.
Instead, we adopt a greedy search strategy over architectural depth and width,
guided by validation feasibility coverage and subset stability.
Once a backbone configuration is selected,
it is fixed across datasets.

Feasibility-related hyperparameters are tuned per dataset. Specifically, we set $\lambda_{\text{close}}=0.50$ for the EMBIC $\leftrightarrow$ NeuroBAT pair
and $\lambda_{\text{close}}=0.45$ for CAS $\leftrightarrow$ SNAP;
all other BFHD hyperparameters are shared across experiments.

In all experiments the training tolerance $\varepsilon_{\text{train}}$ is set to $\varepsilon_{\text{train}} = \tfrac{1}{6}\,\varepsilon_{\text{cutoff}}$.

\subsection{Baseline Implementations}

We include the following baselines.

FRF tune the number of estimators over $\{100,200,500\}$. 

XGBoost models are selected via grid search over $n_{\text{estimators}}\in\{50,60,70,80\}$, $\text{max\_depth}\in\{2,3,4\}$, and $\text{learning\_rate}\in\{0.05,0.1\}$.

The CMLP uses a 4-layer architecture with hidden widths $(64,64,64,32)$, ReLU activations, learning rate $10^{-4}$, and cycle-consistency weight $\lambda_{\text{cycle}}=1$.

The TWMLP replaces the invertible backbone with a non-invertible shared MLP
with hidden widths $(64,64,32)$. TWMLP uses the same output gating mechanism, feasibility losses, annealed $\varepsilon_{\text{train}}$, and learning-rate separation as our approach, isolating the effect of invertibility from that of gating.

\subsection{Protocol for Setting the Feasibility Tolerance $\varepsilon_{\text{cutoff}}$}
\label{sec:sm-tolerance} 
This subsection describes a practical protocol used to instantiate the feasibility tolerance $\varepsilon_{\text{cutoff}}$ in our experiments. It is not a contribution of this work, but a measurement-anchored choice intended to ensure clinical plausibility and reproducibility.
\paragraph{Protocol}
We determine $\varepsilon_{\text{cutoff}}$ based on intrinsic measurement variability of the target assessment system, following the steps below:

\begin{enumerate}
    \item If a subtest reports a standard error of measurement (SEM), we treat this value as an estimate of intrinsic measurement uncertainty in the reported scoring metric.
    \item If SEM is not reported, but test-retest reliability (ICC) and standard deviation (SD) are available, we compute
    \[
    \mathrm{SEM} = \mathrm{SD}\sqrt{1-\mathrm{ICC}}.
    \]
    \item Since reconstruction error in our approach is measured as feature-wise mean squared error in z-scored space, we convert SEM to the corresponding NMSE-scale tolerance via squared normalization:
    \[
    \varepsilon_j
    \;=\;
    \left(\frac{\mathrm{SEM}_j}{\mathrm{SD}_j}\right)^2,
    \]
    where $\mathrm{SD}_j$ denotes the standard deviation of measurement $j$ before normalization.
    \item If $\mathrm{SD}_j$ in the same scoring metric is not available (e.g., SEM is reported on a latent-trait scale or only aggregated uncertainty summaries are provided), we treat the resulting tolerance as a measurement-anchored reference operating point and set $\varepsilon_{\text{cutoff}}$ by a clinically reasonable approximation based on the available information (e.g., conservative upper/lower bounds reported across ability ranges).
    \item We take the largest subtest-level tolerance $\varepsilon_j$ as a conservative system-level feasibility threshold $\varepsilon$.

\end{enumerate}

\paragraph{Usage}
The resulting $\varepsilon_{\text{cutoff}}$ serves as a reference threshold rather than a universal constant. In practice, users may adjust $\varepsilon_{\text{cutoff}}$ within a reasonable range around this baseline depending on application requirements. We therefore report $\varepsilon_{\text{cutoff}}$-sweep analyses to illustrate sensitivity.

\paragraph{Instantiation in this work}
We instantiate the above protocol using measurement statistics reported in prior studies.

\textbf{SNAP-IV.}
For SNAP-IV, test-retest reliability (ICC) and standard deviation (SD) values are reported in \cite{huang2023differences}. Specifically, ICC values of 0.390 (SD = 0.78) for Hyperactivity/Impulsivity are reported. Using $\mathrm{SEM} = \mathrm{SD}\sqrt{1-\mathrm{ICC}}$ and converting to the normalized space used in our experiments, this yields $\varepsilon_{j}$ values of approximately 0.61. We therefore adopt $\varepsilon_{\text{cutoff}} = 0.6$ for SNAP-IV.

\textbf{NeuroBAT (ADNI).}
For NeuroBAT subtests, SEM are directly reported in \cite{crane2023measurement}. As documented in that study, visuospatial function exhibits notably elevated measurement uncertainty. Because these SEM values are reported on a latent ability scale and corresponding feature-level score variances are not available, they cannot be uniquely converted to feature-wise z-space NMSE. We therefore use the reported SEM values to anchor a conservative reference tolerance, and set $\varepsilon_{\text{cutoff}} = 0.6$ for NeuroBAT.

\section{Datasets and Preprocessing}
\label{sec:sm-data}
\subsection{Synthetic Data Generation}
\label{sec:synth}

We construct paired synthetic observations $(X,Y)$ to emulate two heterogeneous clinical assessment systems measured on the same population. The generation process explicitly separates shared (recoverable) structure from system-specific (non-translatable) structure, while incorporating redundancy, weak recoverability, and non-invertible distortions commonly observed in real clinical instruments.

\subsubsection{Latent Variables}

Let
\[
z^{(c)} \in \mathbb{R}^{d_c}, \qquad
z^{(x)} \in \mathbb{R}^{d_x}, \qquad
z^{(y)} \in \mathbb{R}^{d_y}
\]
denote latent random vectors corresponding to shared (common) factors, $X$-specific private factors, and $Y$-specific private factors, respectively. All latent variables are mutually independent and follow standard Gaussian distributions:
\[
z^{(c)} \sim \mathcal{N}(0, I_{d_c}), \quad
z^{(x)} \sim \mathcal{N}(0, I_{d_x}), \quad
z^{(y)} \sim \mathcal{N}(0, I_{d_y}).
\]

\subsubsection{Feature Partition and Ground-Truth Recoverability}

Each system is partitioned into recoverable and non-translatable components:
\[
X = \bigl(X^{\mathrm{rec}}, X^{\mathrm{nt}}\bigr), \qquad
Y = \bigl(Y^{\mathrm{rec}}, Y^{\mathrm{nt}}\bigr).
\]

A feature is labeled recoverable if it depends exclusively on the shared latent variables $z^{(c)}$, up to noise and deterministic transformations. Features depending on private latent variables are labeled non-translatable. Ground-truth recoverability indicators are defined as
\[
\mathrm{GT}_X(j) =
\begin{cases}
1, & \text{if } X_j \text{ depends only on } z^{(c)},\\
0, & \text{otherwise},
\end{cases}
\]
\[
\mathrm{GT}_Y(j) =
\begin{cases}
1, & \text{if } Y_j \text{ depends only on } z^{(c)},\\
0, & \text{otherwise}.
\end{cases}
\]

\subsubsection{Generation of Recoverable Features}

Recoverable features are generated as linear projections of the shared latent variables:
\[
X^{\mathrm{rec}} = z^{(c)} W_X, \qquad
Y^{\mathrm{rec}} = z^{(c)} W_Y,
\]
where $W_X \in \mathbb{R}^{d_c \times k_X}$ and $W_Y \in \mathbb{R}^{d_c \times k_Y}$ are random matrices with i.i.d.\ Gaussian entries.

\paragraph{Redundant recoverable features.}
To model redundant clinical items measuring the same latent construct, we introduce redundancy groups. For each group indexed by $g$, a subset of recoverable features share the same underlying signal:
\[
X^{\mathrm{rec}}_j = X^{\mathrm{rec}}_i, \qquad \forall j \in \mathcal{G}_i,
\]
where $\mathcal{G}_i$ denotes the redundancy group associated with anchor feature $i$. All features within a redundancy group are labeled recoverable in the ground truth.

\paragraph{Weakly recoverable features.}
A subset of recoverable features is designated as weakly recoverable by attenuating their shared signal:
\[
X^{\mathrm{rec}}_j \leftarrow \alpha X^{\mathrm{rec}}_j, \qquad
Y^{\mathrm{rec}}_j \leftarrow \alpha Y^{\mathrm{rec}}_j,
\]
where $0 < \alpha < 1$. These features lie near the $\varepsilon$-recoverability boundary and introduce ambiguity under noise.

\subsubsection{Generation of Non-Translatable Features}

System-specific features are generated exclusively from private latent variables:
\[
X^{\mathrm{nt}} = z^{(x)} A_X, \qquad
Y^{\mathrm{nt}} = z^{(y)} A_Y,
\]
where $A_X \in \mathbb{R}^{d_x \times (d_X - k_X)}$ and $A_Y \in \mathbb{R}^{d_y \times (d_Y - k_Y)}$ are random Gaussian matrices. These features share no information across systems and are therefore non-recoverable by construction.

\subsubsection{Measurement Noise}

Independent Gaussian noise is added to all features:
\[
X^{\mathrm{rec}} \leftarrow X^{\mathrm{rec}} + \varepsilon^{(r)}_X, \qquad
X^{\mathrm{nt}} \leftarrow X^{\mathrm{nt}} + \varepsilon^{(p)}_X,
\]
\[
Y^{\mathrm{rec}} \leftarrow Y^{\mathrm{rec}} + \varepsilon^{(r)}_Y, \qquad
Y^{\mathrm{nt}} \leftarrow Y^{\mathrm{nt}} + \varepsilon^{(p)}_Y,
\]
where $\varepsilon^{(r)} \sim \mathcal{N}(0, \sigma_r^2 I)$ and $\varepsilon^{(p)} \sim \mathcal{N}(0, \sigma_p^2 I)$. The noise levels $\sigma_r$ and $\sigma_p$ control the difficulty of the recoverability task.

\subsubsection{Nonlinear Folding}

To break global invertibility and introduce realistic information loss, each feature undergoes a nonlinear folding operation with probability $p_{\mathrm{nl}}$:
\[
X_j \leftarrow \phi(X_j), \qquad
Y_j \leftarrow \phi(Y_j),
\]
where $\phi(\cdot)$ is randomly chosen from
\[
\phi(u) \in \{ \tanh(u),\; |u|,\; u^2 \}.
\]
These transformations preserve statistical dependence on shared latent variables while destroying strict invertibility.

\subsubsection{Synthetic parameterization (Exp.~1)}
\label{sec:sm-synth-params}

All synthetic results in Exp.~1 use our paired synthetic generator (see released code).

Unless otherwise stated, we use the following parameterization:
\begin{itemize}
    \item Sample size and feature dimensions:
    $n=2000$, $d_X=20$, $d_Y=20$.
    \item Latent dimensions:
    shared latent $d_c=4$; private latents $d_{pX}=4$, $d_{pY}=4$.
    \item Recoverable feature fractions:
    $\texttt{frac\_rec\_x}=0.5$, $\texttt{frac\_rec\_y}=0.5$
    (thus $k_X=10$ and $k_Y=10$ recoverable features per view).
    \item Redundancy in recoverables:
    $\texttt{n\_redundant\_groups\_x}=3$, $\texttt{n\_redundant\_groups\_y}=3$,
    with group size $\texttt{redundant\_group\_size}=3$.
    \item Weak-shared (near-boundary) recoverables:
    $\texttt{weak\_shared\_frac}=0.35$ of recoverable features are attenuated by
    $\texttt{weak\_shared\_scale}=0.30$.
    \item Noise levels:
    recoverable-feature noise $\texttt{noise\_rec}=0.30$;
    private-feature noise $\texttt{noise\_priv}=0.35$.
    \item Nonlinear folding (many-to-one distortion):
    applied independently per feature with probability $\texttt{nonlinear\_prob}=0.95$,
    using $\texttt{fold\_type}=\texttt{mix}$ (randomly chosen from $\tanh(\cdot)$, $|\cdot|$, and $(\cdot)^2$).
\end{itemize}

The generator additionally records (and returns) the indices of recoverable vs.\ non-translatable features,
redundancy groups, weak-recoverable indices, and folded-feature indices, which we use to compute ground-truth
recoverability labels and diagnostics.

\subsubsection{Summary}

This construction yields paired observations $(X,Y)$ in which recoverable and non-translatable features are known by design, yet obscured by redundancy, weak shared structure, noise, and non-invertible distortions. The resulting synthetic datasets enable controlled evaluation of feature-level bidirectional recoverability and stability under realistic conditions.

\subsection{ADNI Cognitive Assessments (EMBIC and NeuroBAT)}

We use paired cognitive assessments from the Alzheimer’s Disease Neuroimaging Initiative (ADNI). Two assessment systems are considered: EMBIC and NeuroBAT.

\paragraph{Assessment construction}
For EMBIC, we use the mean scores of all available tests. Standard deviations are not included, as they are not semantically reconstructable across assessment systems. The resulting EMBIC representation contains 13 numerical measurements.

For NeuroBAT, we include all available subtests, but only use total scores of individual items or subtests.
This design choice avoids incorporating fine-grained ordinal response data. The resulting NeuroBAT representation contains 9 numerical measurements.

\paragraph{Pairing and filtering}
Each subject contributes at most one paired observation. When multiple assessments are available for a subject, only the earliest (first) testing instance is retained. Two assessments from different systems are considered paired if their testing dates fall within a 28-day window; otherwise, they are discarded.

Missing values are handled by complete-case analysis: any sample with missing values in either system is removed. Given the relatively large sample size, no imputation is performed.

\paragraph{Dataset size and availability}
After preprocessing, the ADNI dataset contains 1,629 paired observations. The data are publicly available and can be obtained by application through the ADNI website, subject to standard data use agreements.

\subsection{ADHD Assessments (CAS and SNAP-IV)}

We use a paired ADHD assessment dataset consisting of DN:CAS (CAS) and SNAP-IV (SNAP), provided by our clinical collaborators.

\paragraph{Assessment construction}
For SNAP-IV, we include only the Inattention (ANI) and Hyperactivity/Impulsivity (H/I) subscales. The resulting SNAP-IV representation contains 18 numerical measurements.

For CAS, we use all subscale scores that have been transformed using normative conversion, as well as subtest total scores and the overall scale total score. This yields a CAS representation with 27 numerical measurements.

\paragraph{Pairing and filtering}
As with ADNI, each subject contributes at most one paired observation, corresponding to the earliest available assessment. CAS and SNAP-IV assessments are paired if their testing dates fall within a 28-day window.

Missing values are handled via complete-case analysis: samples with any missing entries in either system are removed. No imputation is applied.

\paragraph{Labels and dataset size}
This dataset includes a binary diagnostic label indicating typical development (TD) or ADHD. The label distribution is balanced. After preprocessing, the dataset contains 336 paired observations.

\paragraph{Data availability}
The data are available upon reasonable request, subject to the relevant data use agreement. For further information on requesting access, please refer to the \href{https://github.com/UOW-AMRL/IJCAI26-BFHD#adhd-dataset} {ADHD dataset section of the project repository}.

\subsection{Data Splitting Protocol}
\label{sec:sm-split}
All datasets are split using a 7:1:2 ratio for training, validation, and testing. Splits are constructed in two stages: first an 8:2 split is performed, followed by a 7:1 split of the training portion into training and validation sets.

To evaluate robustness under temporal shift, all splits for exp.~3 are time-based. Samples are sorted by assessment date, and earlier samples are assigned to training and validation, while later samples are reserved for testing. This protocol is used consistently across all experiments unless otherwise specified.

\section{Additional Utility Analyses}
\label{sec:sm-utility}
All experiments in this section are conducted on the CAS $\leftrightarrow$ SNAP assessment pair, where CAS corresponds to system A ($X$) and SNAP corresponds to system B ($Y$), following the notation used in the main paper.

\subsection{Experiment 5: Downstream Prediction under Size-Matched Harmonized Subsets}
\label{sec:sm-downstream}

This experiment evaluates whether measurement subsets identified as bidirectionally recoverable by different harmonization methods provide practical utility for downstream prediction when only one assessment system is available at deployment time. Unlike Experiments~2-4, which focus on feasibility, coverage, and stability of harmonization itself, this experiment assesses whether harmonized and translated measurements remain usable for an external prediction task. The goal is not to outperform models trained on fully observed target assessments, but to compare methods under a controlled and deployment-relevant setting.

\subsubsection{Experimental Setup}

\paragraph{Dataset and task}
This experiment is conducted only on the CAS (X) $\leftrightarrow$ SNAP (Y) assessment pair. While the main paper evaluates harmonization on multiple assessment systems, downstream prediction in Experiment~5 requires an externally defined label $z$ that is consistently available across paired assessments. Such labels are available for CAS $\leftrightarrow$ SNAP but not for EMBIC $\leftrightarrow$ NeuroBAT, and we therefore restrict this experiment to CAS $\leftrightarrow$ SNAP to avoid introducing task-dependent inconsistencies.

We consider a paired clinical assessment dataset $(X,Y)$ with a single downstream label $z$. The same prediction task and label are used throughout this experiment. Results are reported on one representative real-world dataset due to label availability constraints.

\paragraph{Prediction space and deployment setting}
Prediction is always performed in a fixed feature space, either $X$ or $Y$. At deployment time, only the opposite assessment system is assumed to be available. Accordingly, harmonization models are used to translate the available view into the prediction space before classification. Single-view baselines (X-only and Y-only) use the observed measurements directly in the corresponding prediction space and serve as reference.

\paragraph{Subset construction}
To ensure a fair comparison across methods, we control for subset size. Let $k$ denote the median number of recoverable measurements identified by our approach under the feasibility threshold $\varepsilon_{\text{cutoff}}$ in the prediction space. For our approach, the recoverable subset is obtained by binarizing the learned output gates; if the resulting subset size differs from $k$, measurements are truncated or supplemented based on gate values. For baseline methods without explicit gating, exactly $k$ measurements are selected by ranking measurements according to validation NMSE when predicting the prediction-space view from the opposite view. This yields a size-matched measurement subset for each method.

\paragraph{Prediction models}
For each method and each random seed, a linear SVM classifier is trained independently on the training split using the corresponding feature representation. No classifier weights are shared across methods or seeds. We consider three measurement representations:
\begin{enumerate}
    \item \textbf{Observed:} observed measurements in the prediction space restricted to the selected subset.
    \item \textbf{Translated:} translated measurements in the prediction space, obtained by applying the harmonization model from the opposite view.
    \item \textbf{Single-view reference:} the full observed prediction-space measurements (X-only or Y-only).
\end{enumerate}

To reduce optimistic bias, translated measurements used for training classifiers are generated using a held-out validation split. All classifiers use identical architectures and hyperparameters to isolate the effect of the selected subsets and translation quality.

\paragraph{Evaluation metrics and seeds}
We report classification accuracy (Acc).
Results are reported as mean $\pm$ standard deviation over ten random seeds, using a fixed train/validation/test split.

\subsubsection{Results}

\begin{table}[t]
\centering
\caption{
Downstream prediction under size-matched harmonized subsets. Results are reported as mean $\pm$ standard deviation over ten random seeds. Obs denotes using observed measurements in the prediction space, while Trans denotes using translated measurements obtained from the opposite view. Single-view reference (X-only / Y-only) serve as reference result.
}
\label{tab:sm-downstream}
\vspace{-0.5em}
\begin{tabular}{lcc}
\toprule
\multicolumn{3}{c}{\textbf{Deploy $X\rightarrow \hat{Y}$ (source-side: $X$)}}\\
\midrule
Method & Acc (Obs) & Acc (Trans)\\
\midrule
X-only 
& $0.909 \pm 0.047$ & -  \\

CMLP 
& $0.919 \pm 0.024$ & $0.881 \pm 0.047$ \\

NICE  
& $0.916 \pm 0.023$ & $0.897 \pm 0.033$ \\

Ours  
& $0.913 \pm 0.023$ & $0.901 \pm 0.037$ \\

\midrule
\multicolumn{3}{c}{\textbf{Deploy $Y\rightarrow \hat{X}$ (source-side: $Y$)}}\\
\midrule
Method & Acc (Obs) & Acc (Trans) \\
\midrule
Y-only 
& $0.987 \pm 0.008$ & - \\

CMLP 
& $0.990 \pm 0.010$ & $0.968 \pm 0.018$ \\

NICE  
& $0.987 \pm 0.008$ & $0.974 \pm 0.019$ \\

Ours  
& $0.990 \pm 0.010$ & $0.987 \pm 0.011$ \\

\bottomrule
\end{tabular}
\vspace{-0.5em}
\end{table}

Table~\ref{tab:sm-downstream} summarizes downstream prediction performance under size-matched harmonized subsets. Across both deployment settings, using translated measurements leads to lower classification accuracy than using observed measurements. Nevertheless, BFHD consistently achieves the highest mean accuracy on translated measurements and the smallest performance degradation among the compared harmonization methods.

Overall, these results indicate that feasibility-driven subset identification yields harmonized measurement representations that are not only bidirectionally reconstructable under the chosen tolerance, but also more reliable for downstream prediction when only one assessment system is available at deployment time.

\subsection{Experiment 6: Robustness under MCAR Missingness}
\label{sec:exp6_mcar}

This experiment evaluates the robustness of learned harmonizable measurement subsets under test-time missingness, reflecting deployment scenarios in which one assessment system may be partially incomplete.
Unlike previous experiments, which assume fully observed paired assessments at evaluation time, we introduce missingness only at inference while keeping training conditions unchanged.

\subsubsection{Experimental setup}
All harmonization models are trained using fully observed paired data $(X,Y)$.
At evaluation time, we consider a Missing Completely At Random (MCAR) mechanism applied to one assessment system at a time.
For a missing rate $p \in \{0.1, 0.2, 0.3\}$, each measurement entry in the selected source view is independently dropped with probability $p$.
Missing values are imputed using mean imputation in standardized space, implemented by zero-filling after applying a \texttt{StandardScaler} fit on the training split.
The opposite assessment system remains fully observed.

We evaluate both deployment directions:
(i) missingness in $X$ followed by translation to $Y$ ($X_{\text{miss}} \rightarrow \hat Y$), and
(ii) missingness in $Y$ followed by translation to $X$ ($Y_{\text{miss}} \rightarrow \hat X$).
This symmetric evaluation ensures that robustness is assessed independently of the choice of source assessment.

\subsubsection{Evaluation protocol}
For each method, a reference set of feasible measurements is defined using the clean evaluation setting ($p=0$), which corresponds to the fully observed results reported in Experiment~2. Under increasing missingness, we measure how many of these originally feasible measurements remain feasible.

We report the retention rate, defined as the fraction of originally feasible measurements that remain feasible under missingness. Results are reported as mean $\pm$ standard deviation across random seeds.

\subsubsection{Results and discussion}
\begin{table*}[t]
\centering
\caption{
Robustness under MCAR missingness.
For each method, \#Ret denotes the median number of measurements that remain feasible under missingness. Retain denotes the retention rate relative to the reference feasible set defined at $p=0$ (Experiment~2). Results are reported as mean $\pm$ standard deviation across random seeds.
}
\label{tab:sm-mcar}
\vspace{-0.5em}
\begin{tabular}{c l cc cc cc cc}
\toprule
$p_{\text{miss}}$ & Method 
& \multicolumn{4}{c}{Missing in $X$ ($X_{\text{miss}}\rightarrow \hat Y$)} 
& \multicolumn{4}{c}{Missing in $Y$ ($Y_{\text{miss}}\rightarrow \hat X$)} \\
\cmidrule(lr){3-6} \cmidrule(lr){7-10}
& 
& \#Ret & Retain & Std 
& 
& \#Ret & Retain & Std \\
\midrule
0.1
& Ours  
& $9$   & $0.860$ & $0.097$ 
& 
& $7$   & $0.813$ & $0.140$ \\

& NICE  
& $8.5$ & $0.788$ & $0.152$ 
& 
& $7$   & $0.877$ & $0.121$ \\

& CMLP
& $8$   & $0.815$ & $0.196$ 
& 
& $6$   & $0.725$ & $0.293$ \\
\midrule
0.2
& Ours  
& $6.5$ & $0.513$ & $0.158$ 
& 
& $5.5$ & $0.634$ & $0.112$ \\

& NICE  
& $6.5$ & $0.524$ & $0.203$ 
& 
& $6$   & $0.632$ & $0.168$ \\

& CMLP
& $6$   & $0.676$ & $0.277$ 
& 
& $4$   & $0.579$ & $0.317$ \\
\midrule
0.3
& Ours  
& $4.5$ & $0.333$ & $0.142$ 
& 
& $2.5$ & $0.315$ & $0.142$ \\

& NICE  
& $2.5$ & $0.240$ & $0.145$ 
& 
& $2$   & $0.286$ & $0.130$ \\

& CMLP
& $2$   & $0.226$ & $0.143$ 
& 
& $2$   & $0.296$ & $0.269$ \\
\bottomrule
\end{tabular}
\vspace{-0.5em}
\end{table*}

Table~\ref{tab:sm-mcar} summarizes robustness under increasing levels of test-time missingness.
Across both deployment directions, retention rates decrease monotonically as the missing rate increases, reflecting the expected degradation under partial input corruption.

At mild missingness ($p_{\text{miss}}=0.1$), our approach consistently retains a larger fraction of originally feasible measurements than CMLP and is competitive with NICE, while exhibiting lower variability across seeds. At moderate and severe missingness levels ($p_{\text{miss}}=0.2$ and $0.3$), ours maintains comparable or higher retention rates than baselines in both directions, with more stable behavior under increasing uncertainty.

Importantly, robustness trends are consistent across the two deployment directions, indicating that the learned bidirectional structure is not overly sensitive to which assessment system is partially observed at inference time.
These results suggest that feasibility-driven measurement decomposition yields harmonization decisions that degrade more gracefully under test-time missingness, supporting their use in realistic deployment settings where assessments may be incomplete.

\section{Extended Experimental Results}
\label{sec:sm-results}
\subsection{Additional Results for Experiment 1: Synthetic Experiments}
\label{sec:sm-exp1}
\begin{table*}[t]
\centering
\caption{
Experiment~1 (Synthetic): feasibility structure recovery and stability across different data-generation seeds ($s_{\mathrm{gen}} \in \{1,\dots,5\}$). Acc is index-level recoverability label accuracy (mean $\pm$ std over training seeds). $J_X$ and $J_Y$ are Jaccard similarities of selected recoverable subsets across training seeds; $J_{\mathrm{overall}}$ is the average of $J_X$ and $J_Y$. CCC statistics summarize agreement strength among selected measurement indices (mean ± std).
}
\label{tab:sm-exp1-full}
\vspace{-0.5em}
\begin{tabular}{l l c c c c c c c c}
\toprule
Data & Method 
& Acc$_X$ 
& Acc$_Y$ 
& $J_X$ 
& $J_Y$
& $J_{\mathrm{overall}}$
& CCC$_X$
& CCC$_Y$ \\
\midrule
& Ours 
& $0.915 \pm 0.023$ 
& $0.900 \pm 0.000$ 
& 0.948 & 1.000 & 0.974
& $0.738 \pm 0.014$
& $0.754 \pm 0.063$ \\
$s_{\mathrm{gen}}$ = 1& CMLP 
& $0.890 \pm 0.037$ 
& $0.900 \pm 0.000$ 
& 0.896 & 1.000 & 0.948
& $0.768 \pm 0.019$
& $0.758 \pm 0.014$ \\
& NICE 
& $0.850 \pm 0.000$ 
& $0.805 \pm 0.047$ 
& 1.000 & 0.824 & 0.912
& $0.722 \pm 0.013$
& $0.745 \pm 0.013$ \\
\midrule
& Ours 
& $0.830 \pm 0.056$ 
& $0.755 \pm 0.042$ 
& 0.783 & 0.812 & 0.797
& $0.574 \pm 0.033$
& $0.619 \pm 0.049$ \\
$s_{\mathrm{gen}}$ = 2& CMLP 
& $0.810 \pm 0.037$ 
& $0.710 \pm 0.030$ 
& 0.848 & 0.722 & 0.785
& $0.612 \pm 0.023$
& $0.678 \pm 0.024$ \\
& NICE 
& $0.550 \pm 0.032$ 
& $0.605 \pm 0.035$ 
& 0.400 & 0.544 & 0.472
& $0.668 \pm 0.043$
& $0.648 \pm 0.016$ \\
\midrule
& Ours 
& $0.895 \pm 0.015$ 
& $0.885 \pm 0.050$ 
& 0.975 & 0.858 & 0.916
& $0.622 \pm 0.010$
& $0.616 \pm 0.028$ \\
$s_{\mathrm{gen}}$ = 3& CMLP 
& $0.845 \pm 0.052$ 
& $0.825 \pm 0.025$ 
& 0.773 & 0.846 & 0.809
& $0.659 \pm 0.019$
& $0.683 \pm 0.024$ \\
& NICE 
& $0.615 \pm 0.032$ 
& $0.650 \pm 0.032$ 
& 0.624 & 0.800 & 0.712
& $0.741 \pm 0.023$
& $0.740 \pm 0.023$ \\
\midrule
& Ours 
& $0.840 \pm 0.020$ 
& $0.875 \pm 0.025$ 
& 0.883 & 0.931 & 0.907
& $0.576 \pm 0.028$
& $0.595 \pm 0.015$ \\
$s_{\mathrm{gen}}$ = 4& CMLP 
& $0.745 \pm 0.052$ 
& $0.855 \pm 0.047$ 
& 0.776 & 0.815 & 0.796
& $0.648 \pm 0.024$
& $0.622 \pm 0.023$ \\
& NICE 
& $0.590 \pm 0.030$ 
& $0.620 \pm 0.024$ 
& 0.569 & 0.822 & 0.695
& $0.728 \pm 0.048$
& $0.703 \pm 0.026$ \\
\midrule
& Ours 
& $0.880 \pm 0.024$ 
& $0.830 \pm 0.024$ 
& 0.933 & 0.924 & 0.929
& $0.706 \pm 0.029$
& $0.690 \pm 0.025$ \\
$s_{\mathrm{gen}}$ = 5& CMLP 
& $0.855 \pm 0.015$ 
& $0.810 \pm 0.049$ 
& 0.975 & 0.789 & 0.882
& $0.728 \pm 0.018$
& $0.716 \pm 0.035$ \\
& NICE 
& $0.735 \pm 0.039$ 
& $0.725 \pm 0.040$ 
& 0.837 & 0.814 & 0.826
& $0.698 \pm 0.024$
& $0.683 \pm 0.015$ \\
\bottomrule
\end{tabular}
\vspace{-0.5em}
\end{table*}

\paragraph{Remarks}
Table~\ref{tab:sm-exp1-full} summarizes index-level structure recovery results across different data-generation seeds. Across all settings, BFHD achieves consistently strong recoverability accuracy and high subset stability. Stability is better reflected by Jaccard similarity ($J_X$, $J_Y$) than by accuracy variance, since index-level accuracy is a discrete metric that can saturate once the correct subset size is reliably identified.

CCC statistics are reported as a diagnostic measure of agreement strength among selected measurement indices. We note that the BFHD objective is feasibility-driven rather than accuracy-driven: the goal is to identify large bidirectionally recoverable subsets under a prescribed tolerance, not to optimize pointwise reconstruction or agreement metrics. Accordingly, CCC values need not be maximal, and moderate CCC levels are expected for measurements that lie near the feasibility boundary.

\subsection{Additional Results for Experiment 2: Real-World Harmonization}
\label{sec:sm-exp2}

This section provides supplementary results for Experiment~2 on the CAS$\leftrightarrow$SNAP assessment pair.
While the main paper reports fixed-threshold harmonization outcomes (including subset stability diagnostics) and presents an $\varepsilon$-sweep for EMBIC$\leftrightarrow$NeuroBAT, the analysis below complements the main text by showing the corresponding $\varepsilon$-sweep on CAS$\leftrightarrow$SNAP.

\begin{figure*}[t]
    \centering
    \includegraphics[width=0.8\linewidth]{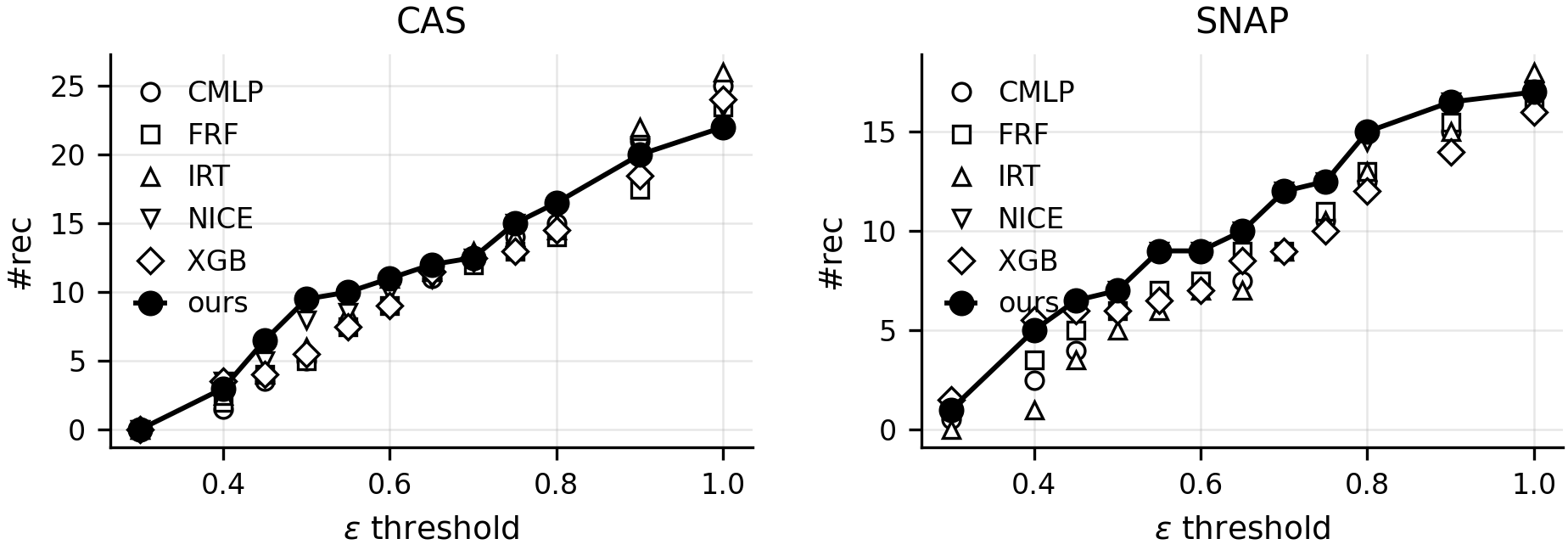}
    \caption{
    \textbf{$\varepsilon$-sweep on CAS$\leftrightarrow$SNAP.}
    Median number of recoverable measurements (\#Rec) as a function of the feasibility threshold $\varepsilon$, shown separately for CAS (A) and SNAP (B).
    Results are aggregated over random seeds.
    }
    \label{fig:sm-epsilon-sweep-cas-snap}
\end{figure*}

\paragraph{Feasibility-coverage trends on CAS$\leftrightarrow$SNAP}
Figure~\ref{fig:sm-epsilon-sweep-cas-snap} reports the median number of recoverable measurements (\#Rec) as a function of the feasibility threshold $\varepsilon$, shown separately in the CAS (A) and SNAP (B) spaces.
All methods are evaluated under the same feasibility criterion used throughout Experiment~2.

Across a broad and clinically relevant range of thresholds, BFHD consistently identifies a larger number of feasible measurements than baseline methods in both directions.
The advantage is most pronounced in the intermediate regime, where partial but nontrivial overlap between the two assessment systems is expected.
As $\varepsilon$ increases, all methods approach full coverage; however, BFHD exhibits an earlier and smoother increase in \#Rec across both assessment spaces, yielding a more interpretable coverage-feasibility frontier.

\paragraph{Remark}
Subset stability across random seeds is reported in the main paper (via Jaccard similarity) at the reference threshold, and the CAS$\leftrightarrow$SNAP $\varepsilon$-sweep shown here is consistent with the same qualitative behavior: BFHD expands feasible harmonization coverage under a shared feasibility criterion.

\subsection{Additional Results for Experiment 3: Temporal Deployment under Protocol Drift}
\label{sec:sm-exp3}
This section reports additional temporal deployment results under protocol drift for both real-world assessment pairs.
For completeness and ease of comparison, we include the EMBIC$\leftrightarrow$NeuroBAT results shown in the main paper and extend them with agreement diagnostics on the held-out test period.
In addition, we report the corresponding results on CAS$\leftrightarrow$SNAP, which are omitted from the main paper due to space constraints.

\paragraph{Protocol and metrics.}
The experimental protocol is identical to that described in the main paper.
For each deployment direction (source $\rightarrow$ target), we report:
(i) \#Ret, the median number of training-period recoverable target-space measurements that remain feasible on the test period;
(ii) Ret\%, the retention rate on the test period (mean $\pm$ std across random seeds);
and (iii) CCC-based diagnostic counts computed over retained measurements on the test period, reported as median counts across seeds.

\begin{table*}[t]
\centering
\caption{
Extended results for experiment 3.
For each deployment direction (source $\rightarrow$ target), \#Ret is the median number of retained recoverable measurements in the target space, and Ret\% is the corresponding retention rate (mean $\pm$ std across random seeds), and is reported in percentage units.
CCC$>$0.5/0.6/0.7 report the median number of retained measurements whose concordance correlation coefficient on the test period exceeds the corresponding threshold.
``-'' denotes not applicable (no recoverable measurement is identified on the training period, hence retention is undefined).
All results use $\varepsilon_{\text{cutoff}}=0.6$.
}
\label{tab:sm-exp3-temporal-full}
\vspace{-0.5em}
\begin{tabular}{l l r c r r r}
\toprule
Direction & Method & \#Ret & Ret\% (mean$\pm$std) & CCC$>$0.5 & CCC$>$0.6 & CCC$>$0.7 \\
\midrule
\multicolumn{7}{l}{\textbf{EMBIC $\rightarrow$ NeuroBAT (target: NeuroBAT)}}\\
FRF   &  & 0   & $0.0\pm0.0$ & 0 & 0 & 0 \\
IRT   &  & 0   & -              & 0 & 0 & 0 \\
XGB   &  & 0.5 & $62.5\pm48.4$ & 0.5 & 0 & 0 \\
CMLP  &  & 1   & $90.0\pm10.0$ & 1 & 1 & 0 \\
NICE  &  & 2   & $49.2\pm10.5$ & 2 & 1 & 0 \\
Ours  &  & 3   & $100.0\pm0.0$ & 3 & 1 & 0 \\
\midrule
\multicolumn{7}{l}{\textbf{NeuroBAT $\rightarrow$ EMBIC (target: EMBIC)}}\\
FRF   &  & 5 & $56.0\pm8.0$ & 5 & 5 & 2 \\
IRT   &  & 3 & $42.9\pm0.0$ & 3 & 3 & 2 \\
XGB   &  & 6 & $85.7\pm0.0$ & 6 & 6 & 3 \\
CMLP  &  & 6 & $71.0\pm7.5$ & 6 & 6 & 3 \\
NICE  &  & 6 & $77.4\pm9.9$ & 6 & 6 & 3 \\
Ours  &  & 7 & $100.0\pm0.0$ & 7 & 6.5 & 3 \\
\midrule
\multicolumn{7}{l}{\textbf{CAS $\rightarrow$ SNAP (target: SNAP)}}\\
FRF   &  & 7   & $71.4\pm23.9$ & 7 & 7 & 5 \\
IRT   &  & 5.5 & $56.1\pm38.0$ & 5.5 & 5.5 & 2.5 \\
XGB   &  & 7   & $71.4\pm13.3$ & 7 & 7 & 6 \\
CMLP  &  & 7   & $74.0\pm17.9$ & 7 & 7 & 5.5 \\
NICE  &  & 9   & $79.4\pm12.4$ & 9 & 7.5 & 5.5 \\
Ours  &  & 9   & $93.8\pm9.0$ & 9 & 7 & 5.5 \\
\midrule
\multicolumn{7}{l}{\textbf{SNAP $\rightarrow$ CAS (target: CAS)}}\\
FRF   &  & 8   & $64.5\pm30.9$ & 8 & 8 & 5 \\
IRT   &  & 6.5 & $45.8\pm36.8$ & 6.5 & 6 & 4.5 \\
XGB   &  & 8   & $63.5\pm36.8$ & 8 & 8 & 5 \\
CMLP  &  & 8.5 & $73.0\pm17.4$ & 8.5 & 8 & 4 \\
NICE  &  & 10  & $67.8\pm17.3$ & 10 & 7 & 2.5 \\
Ours  &  & 11  & $86.3\pm14.9$ & 11 & 6.5 & 4 \\
\bottomrule
\end{tabular}
\vspace{-0.8em}
\end{table*}

\paragraph{Remarks}
Across both dataset pairs and both deployment directions, BFHD retains a larger fraction of training-period recoverable measurements on the later cohort while maintaining strong or comparable CCC-based agreement diagnostics among the retained measurements.
Notably, for EMBIC$\rightarrow$NeuroBAT, several baselines identify few or no training-period recoverable target measurements, resulting in undefined retention rates; BFHD avoids this failure mode and yields stable retention across random seeds.

\subsection{Additional Results for Experiment 4: Ablation Results on CAS$\leftrightarrow$SNAP}
\label{sec:sm-exp4-cas}

This section reports additional ablation results for Experiment~4 on the CAS$\leftrightarrow$SNAP assessment pair.
The ablation settings, feasibility threshold, and evaluation metrics are identical to those used in the main paper.
These results are included to verify that the qualitative conclusions drawn from EMBIC$\leftrightarrow$NeuroBAT generalize to a second real-world dataset with different measurements dimensionality and assessment structure.

Table~\ref{tab:sm-exp4-ablation-cas-snap} summarizes recoverable subset size and stability across random seeds.
Consistent with the main paper, replacing the invertible backbone with a tied-weight MLP (TWMLP) yields comparable recoverable coverage and subset stability.
In contrast, removing output gating leads to noticeably reduced stability, despite similar coverage.
Replacing the gate reward with a standard $L_0$ penalty or removing the hinge-based feasibility loss results in degenerate gating behavior and therefore does not admit meaningful stability statistics.

\begin{table}[t]
\centering
\caption{
Ablation results on CAS (A) $\leftrightarrow$ SNAP (B) under $\varepsilon_{\text{cutoff}}=0.6$.
\#Rec$_A$/\#Rec$_B$ denote median numbers of recoverable measurements. $J_A$/$J_B$ denote Jaccard similarity across random seeds.
Degenerate variants are marked with ``-''.
}
\label{tab:sm-exp4-ablation-cas-snap}
\vspace{-0.5em}
\begin{tabular}{l c c c c}
\toprule
Variant 
& $\#\mathrm{Rec}_A$
& $J_A$
& $\#\mathrm{Rec}_B$
& $J_B$\\
\midrule
Ours (full)          & 11 & 0.930 & 9 & 0.817 \\
\midrule
TWMLP backbone   & 11 & 0.929 & 9 & 0.947 \\
$L_{0}$ penalty      & $0^{\dagger}$ & - & $0^{\dagger}$ & - \\
No gating        & 11 & 0.806 & 9 & 0.774 \\
No hinge         & $27^{\dagger}$ & - & $18^{\dagger}$ & - \\
\bottomrule
\end{tabular}
\begin{flushleft}
{\footnotesize
$\dagger$ Trivial gating solutions. $L_0$ penalty collapses to all-closed gating, while removing the hinge leads to all-open or unstable collapse depending on hyperparameter settings. Subset stability is not meaningful for these variants.
}
\end{flushleft}
\end{table}

\clearpage

\section*{Ethical Statement}
This study uses de-identified datasets, including publicly available datasets and access-controlled datasets available upon reasonable request under appropriate approvals. No new data collection or human subject recruitment was conducted.

\section*{Acknowledgments}
Data collection and sharing for the Alzheimer’s Disease Neuroimaging Initiative (ADNI) is funded by the National Institute on Aging, National Institutes of Health Grant U19 AG024904. The grantee organization is the Northern California Institute for Research and Education.

\section*{Contribution Statement}
Yuanhao Zhuo and Zixi Qin contributed equally to this work.

\bibliographystyle{named}
\bibliography{ijcai26}
\end{document}